\documentclass[conference]{IEEEtran}
\IEEEoverridecommandlockouts
\usepackage{cite}
\usepackage{amsmath,amssymb,amsfonts}
\usepackage{algorithmic}
\usepackage{graphicx}
\usepackage{textcomp}
\usepackage{xcolor}
\usepackage[a4paper, total={184mm,239mm}]{geometry}
\def\BibTeX{{\rm B\kern-.05em{\sc i\kern-.025em b}\kern-.08em
    T\kern-.1667em\lower.7ex\hbox{E}\kern-.125emX}}
\begin{document}

\title{Biologically Plausible Learning on Neuromorphic Hardware Architectures}

\author{\IEEEauthorblockN{Christopher Wolters\textsuperscript{1,2},
Brady Taylor\textsuperscript{1}, Edward Hanson\textsuperscript{1}, Xiaoxuan Yang\textsuperscript{1}, Ulf Schlichtmann\textsuperscript{2} and Yiran Chen\textsuperscript{1}}
\IEEEauthorblockA{\textsuperscript{1}Department of Electrical and Computer Engineering,
Duke University, Durham, NC\\
\textsuperscript{2}Institute for Electronic Design Automation, Technical University of Munich, Munich, Germany \\
Email: \textsuperscript{1}\{brady.g.taylor, edward.t.hanson, xiaoxuan.yang1, yiran.chen\}@duke.edu, \textsuperscript{2}\{ch.wolters, ulf.schlichtmann\}@tum.de}}

\maketitle
\begin{abstract}
With an ever-growing number of parameters defining increasingly complex networks, Deep Learning has led to several breakthroughs surpassing human performance. As a result, data movement for these millions of model parameters causes a growing imbalance known as the memory wall. Neuromorphic computing is an emerging paradigm that confronts this imbalance by performing computations directly in analog memories. On the software side, the sequential Backpropagation algorithm prevents efficient parallelization and thus fast convergence. A novel method, Direct Feedback Alignment, resolves inherent layer dependencies by directly passing the error from the output to each layer. At the intersection of hardware/software co-design, there is a demand for developing algorithms that are tolerable to hardware nonidealities. Therefore, this work explores the interrelationship of implementing bio-plausible learning in-situ on neuromorphic hardware, emphasizing energy, area, and latency constraints. Using the benchmarking framework DNN+NeuroSim, we investigate the impact of hardware nonidealities and quantization on algorithm performance, as well as how network topologies and algorithm-level design choices can scale latency, energy and area consumption of a chip. To the best of our knowledge, this work is the first to compare the impact of different learning algorithms on Compute-In-Memory-based hardware and vice versa. The best results achieved for accuracy remain Backpropagation-based, notably when facing hardware imperfections. Direct Feedback Alignment, on the other hand, allows for significant speedup due to parallelization, reducing training time by a factor approaching N for N-layered networks.
\end{abstract}


\section{Introduction}
As model complexities in deep neural networks (DNN) increase, machine learning programs continue to achieve remarkable successes, even surpassing human performance~\cite{b1}. These models place heavy demands on computing power, a problem exacerbated by the discontinuation of Moore's Law, as the limits of transistor technology are being reached~\cite{b2}. Fetching, computing, and updating millions of weights puts enormous demand on data movement~\cite{b3}. In particular, a growing imbalance results from the strict separation of computation and memory in von Neumann architectures. Expensive memory accesses are communicated via buses with limited bandwidth, creating a drastic bottleneck: two orders of magnitude more energy is consumed by data transfers between memory and chip than by floating-point operations (FLOPs)~\cite{b2}.

Computation-in-Memory (CIM) is an emerging solution in which memory modules and data processing units are unified to reduce data transportation from memory to chip. Resistive random-access memory (RRAM) is a new generation of non-volatile memory systems with low access time and energy consumption~\cite{b5}. The ability of RRAM macros to execute high-throughput dot-products with high computational density and energy efficiency is desirable for constructing real-time, power-constrained computing systems~\cite{b3}. Therefore, this work focuses specifically on the interrelationship between deep learning and neuromorphic hardware, with an emphasis on energy, area, and latency.

While the brain has always inspired computing paradigms, there is little empirical relation between deep learning and neuroscience. In fact, the common Backpropagation (BP) algorithm is often criticized for its biological implausibility~\cite{b6}. Launay et al.~\cite{b7} primarily cite the global dependence of each of the network's layers on postsynaptic layers. Known as ``backward locking'', this prevents the network from parallelizing the learning process. Furthermore, biological synapses are assumed to be unidirectional and do not allow symmetrical weight structures in feedforward and feedback directions, commonly cited as the ``weight transportation problem''~\cite{b8}. Consequently, new algorithms have been developed with more faithful neural behavior. A family of related approaches was introduced by N{\o}kland~\cite{b9}, including Direct Feedback Alignment (DFA), which successfully solves these implausibilities and achieves competitive performance despite a fundamentally altered computational flow~\cite{b9}.

Furthermore, there is need for algorithms that can robustly tolerate the nonideal properties of neuromorphic hardware and fully seize their potential as a novel computing scheme. In this work, we set out to analyze different learning rules' performance on neuromorphic hardware architectures and vice-versa. This involves measuring the effects of in-memory implementations, with particular attention to chip area, latency, and energy, while improving the software-side accuracy. The primary focus is on deep learning and its hardware/software co-design. Instead of optimizing absolute learning performance, relative differences will be compared over different technologies and various metrics. In Section II, we will provide a more detailed description of the DFA algorithm and CIM schemes. In Section III, we will outline our approach for determining algorithm effects on hardware and vice-versa. Finally, we will discuss our results in Section IV and conclude the paper in Section V.
\section{Technical Background}
\subsection{Direct Feedback Alignment}
The central question of credit assignment in neural networks is answered in backpropagation by passing back the error as a teaching signal. Mathematically, this corresponds to calculating the gradient with respect to the output of each layer, leading to (\ref{backprop}), where $\odot$ denotes the elementwise Hadamard product, $W$ is the $i$th layer's weight matrix, $a$ is its activation. Further $\delta a_N$ is defined as the error, $e$.
\begin{equation} \label{backprop}
    \forall i \in [1, ..., N-1]: \quad \delta a_i = \frac{\partial L}{\partial a_i} = (W_{i+1}^T \delta a_{i+1}) \odot f_i'(a_i)
\end{equation}
The final weight updates are given as 
\begin{equation}
    \forall i \in [1, ..., N]: \quad\delta W_i = -\delta a_i  h_{i-1}^T
\end{equation}
Parameter updates are propagated backward sequentially through the network, allowing ``blame'' for the error to be assigned precisely to each neuron. This structure results in a computational flow where forward and backward passes are performed on the same set of weights. The update of synaptic weights requires all information of the forward process to be known within the backward process, increasing buffer requirements in hardware. Furthermore, each neuron needs precise knowledge about neurons of subsequent layers, making the gradient calculation global. In contrast, the brain is highly parallel, asymmetric, and local in its computations. However, BP is unable to parallelize the backward pass due to its sequential layer dependencies, leading to a training bottleneck within deep networks.

One potential solution to the biological implausibilities and resulting performance limitations of BP is Direct Feedback Alignment (DFA), introduced by N{\o}kland~\cite{b9}. While the algorithm retains the same general structure of BP, it employs independent forward and backward paths. Instead of symmetric weights, random matrices are incorporated for the error projection. Still, learning occurs as the network learns how to make the teaching signal useful. In mathematical terms, Equation~(\ref{dfaeq}) defines the error projection for DFA, whereas the formulas for updating the weights and the last layer's error remain the same.
\begin{equation} \label{dfaeq}
    \forall i \in [1, ..., N-1]: \quad \delta a_i = (B_ie) \odot f_i'(a_i)
\end{equation}
Compared to (\ref{backprop}), the fixed random matrix $B_i$, with corresponding dimensions \emph{output features} $\times$ \emph{number of classes}, replaces the transposed forward weight matrix, $W_i^T$. Furthermore, the error $e$ is passed directly to each layer in parallel in place of $\delta a_{i+1}$, resolving layer dependencies and making it an almost entirely local and biologically plausible process~\cite{b9}. Experiments by N{\o}kland~\cite{b9}, Launay et al.~\cite{b12}, and Sanfiz et al.~\cite{b8} show DFA's performance on par with BP for simple tasks like MNIST \cite{b15} image classification to state-of-the-art methods such as Graph Convolutional Networks. Yet, DFA fails to train Convolutional Neural Networks, as experiments by Launay et al.\cite{b12} show. Hence, this work focuses on training fully connected networks only.

\subsection{Computing-in-Memory}
Despite fundamental differences in the computational process, both learning rules are based on matrix-vector multiplications, which require large numbers of parallel MAC (Multiply \& Accumulate) operations, as seen in Equations (\ref{backprop}) and (\ref{dfaeq}). The number of computations required increases with network depth, and therefore, the demand for computational power on underlying hardware systems grows extensively. CIM aims to close the gap between memory and computation in the von Neumann model: By storing the network's weights in memory, the analog nature of the underlying devices can be exploited to perform MAC operations and directly incorporate the physical memory in computation. Since the weight transfer from memory to chip is no longer needed, high performance and energy efficiency are achieved~\cite{b13}. In addition, similar to the brain, the calculations are carried out in a highly parallelized scheme, allowing for further computational speedup.

The MAC operations are executed in analog memory by encoding the input vector as a voltage signal $V$ and applying to the memory cells, which are programmed to conductances representing matrix weights, along each row. According to Ohm's Law, the current through each device is the product of $V$ and the device conductance, while currents accumulate down each column according to Kirchhoff's current law. Hence, the resulting currents
\begin{equation}
    I_j = \sum_{i=1}^{n} V_i g_{ij}
\end{equation}
are summed along the column, where $g_{ij}$ is the conductance of the memristor in the $i$th row and $j$th column. This corresponds to a dot product of the column conductances and the input voltage signal. Each column of weights can represent a separate dot product executed in parallel. Results can be obtained with a single read operation, enabling enormous throughput and high energy efficiency~\cite{b13}.

However, this scheme encounters key challenges involving various device- and circuit-level nonidealities attributed to the memory cells and data-conversion circuitry. Potential issues arise from both the fabrication and computing processes; e.g., process variations cause devices across a single chip to behave differently regarding conductivity ranges and programmability, and analog-to-digital converters (ADCs) have limited precision in interpreting analog outputs from memory. The resulting errors accumulate in matrix-vector multiplications and degrade the model's performance. It is therefore critical to study the impact of hardware-induced imperfections on large-scale DNNs. 
\section{Approach}
Due to these nonidealities, tailored solutions are required for both hardware and software. To address these tasks, Peng et al.~\cite{b10} introduced DNN+NeuroSim as a circuit-level macro model for benchmarking neuro-inspired architectures on performance metrics such as chip area, latency, and energy. It supports a hierarchical organization from the device level (memory technology) through the circuit level (peripherals) and chip level (tiles, processing elements \& subarrays) to the algorithm level (neural network topologies)~\cite{b10}. 

\begin{figure}[tb]
    \centering
    \includegraphics[width=0.49\textwidth]{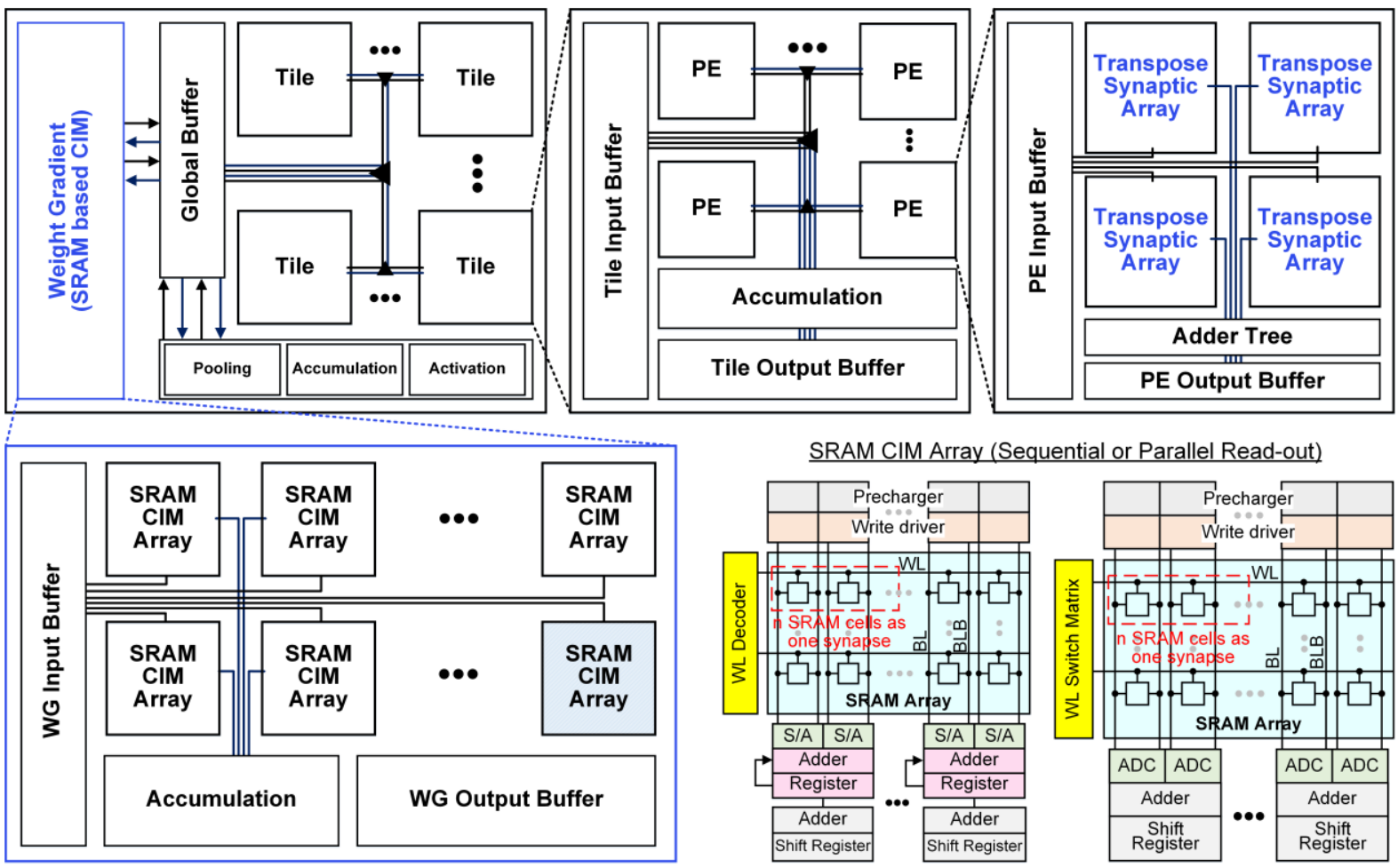}
    \caption{Schematic of chip floorplan for NeuroSim~\cite{b10}. The hierarchical design can be subdivided by chip into tiles, which are comprised of processing elements (PEs). These PEs are further divided into synaptic arrays of many possible memory types.}
    \label{fig:neurosim}
\end{figure}

To model different technologies and levels of abstraction, the structure of the underlying hardware scheme is shown in Fig.~\ref{fig:neurosim}. In general, it is assumed that the on-chip memory is sufficient to store the synaptic weights of the entire neural network. Thus, the only access to off-chip memory is to fetch the input data and buffer intermediate values. From a top-down perspective, the chip consists of multiple tiles on which the weights are stored, a global buffer, and neural function units, including accumulation and activation units. Each tile can be further broken down into several processing elements, a tile buffer for loading activations, accumulation modules for partial sums, and an output buffer. At the lowest level, a processing element is composed of subarrays, which perform the analog calculations. For computing weight gradients, special Weight Gradient Units (WGUs) on static random-access memory (SRAM) are utilized~\cite{b10}.

The simulator must be adapted to fit the computational requirements of DFA. A direct consequence of the additional backward matrices introduced in DFA includes additional memory requirements. Launay et al.~\cite{b7}, however, propose using sliced matrices from the matrix with the largest dimensions. As all backward matrices agree on the error dimension, the second dimension can be chosen as the highest number of output features. Nevertheless, to compute the gradients of $N$ layers in parallel, $N$ WGUs are required. In contrast, only one is needed for sequential Backpropagation.

Another implication of using random matrices for error feedback as opposed to transposed weight matrices is the elimination of the need for transposable subarrays. To support on-chip training with BP, synaptic arrays must be modified to support forward and backward computations: a transposable readout scheme realizes transposed computations with additional peripheral circuitry. Decoders, shift adders, and ADCs are duplicated and rotated 90° from the original~\cite{b10}. Without the need for transposable arrays, the duplicated peripheral circuitry is no longer needed.

One more architectural difference between BP and DFA results from the different dimensions of the matrices used during the backward pass. As the error dimension used for DFA is typically smaller than the number of hidden features in BP, and as the FLOPs for matrix-multiplications scale proportionally to their dimension, DFA consumes potentially less computational energy. Having outlined potential sources for differences in our design metrics of accuracy, area, latency, and energy, a multiobjective optimization problem arises. Different functions for these metrics need to be maximized or minimized, and ideal approaches should be found to optimize each target value while considering their impact on other metrics. 
\section{Discussion}
With previous works by N{\o}kland~\cite{b9}, Launay et al.~\cite{b12}, and Sanfiz et al.~\cite{b8} showing DFA's accuracy on par with BP's under given constraints, the focus of the following experiments is predominantly on the relative hardware comparisons between the algorithms and their scaling in the context of deep learning. As the estimates were relatively constant over the training process, only the breakdown of the first epoch is depicted. Throughout the experiments, a default network of five layers with 1024 hidden features per layer is favored as BP and DFA achieve comparable classification performance on the Fashion MNIST~\cite{b14} task with this topology. While different analyses might involve adjustments to these parameters, any changes will be indicated.

\subsection{Model Performance of Direct Feedback Alignment}
\begin{figure}[h!]
    \centering
    \includegraphics[width=0.49\textwidth]{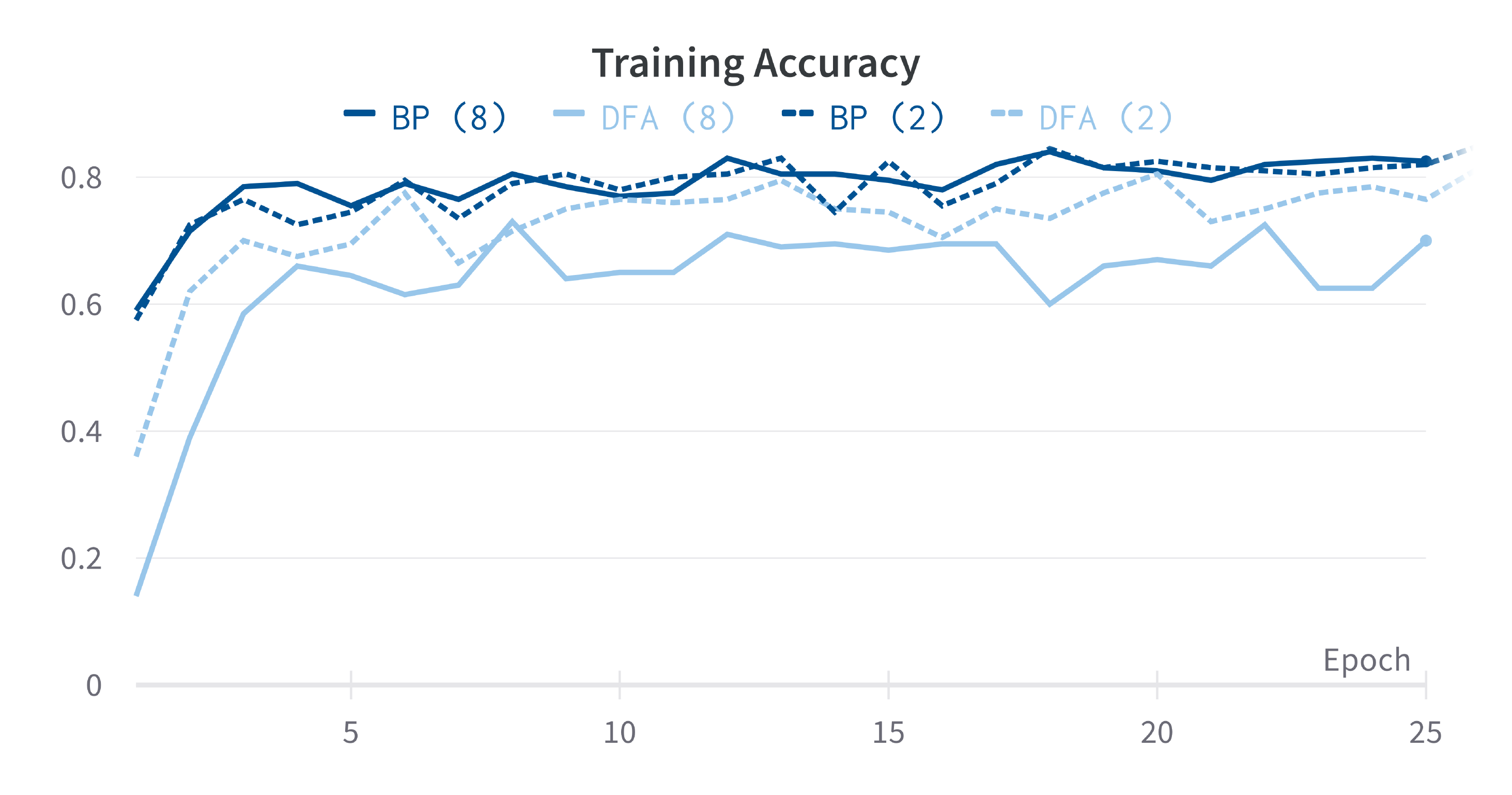}
    \caption{Accuracy of DFA and BP over 25 epochs for 2 layers and 8 layers. These results demonstrate that DFA requires more epochs for convergence as network depth increases.}
    \label{fig:howmany}
\end{figure}

When sweeping different hardware and network hyperparameters, it is vital to consider the general performance of DFA compared to BP. This ensures comparable performance on the classification task, which remains the fundamental output measure. Experiments by Sanfiz et al.~\cite{b8} and N{\o}kland~\cite{b9} indicate that the best performance for DFA is obtained by using Xavier initialization and ReLu activation functions. Furthermore, we observed substantially better results when training with larger batches, seemingly averaging out the noise induced by DFA's inherent gradient approximation. Initially, we compared the number of iterations required for the models to converge to a stable training accuracy. Fig.~\ref{fig:howmany} displays the resulting model performance for each learning rule on a 2-layered network compared to 8 layers.

This leads to an important insight for latency evaluations within deep networks: as the network depth increases, DFA requires more epochs to converge. Especially within the first few epochs, worse performance is observed, whereas BP appears slightly more stable in its convergence. Refinetti et al.~\cite{b11} characterize this phenomenon as ``memorise, then learn'', where first the matrices $W$ and $B$ are aligned before the learning begins. Since this process of alignment occurs sequentially, the effect becomes more noticeable for deeper networks. A second drawback of the required alignment phase becomes more apparent when different matrix widths, i.e., the number of hidden features, are swept. Table~\ref{tab:width} compares the final test accuracy for both learning rules after 100 epochs of training over different matrix widths. 

\begin{table}[h]
    \centering
    \caption{Final test accuracy for different layer widths}
    \begin{tabular}{c | c | c}
    \textbf{Width} &\textbf{DFA} &\textbf{BP} \\
    \hline
        8 &0.08 &0.44\\
        16 &0.53 &0.73\\
        32 &0.69 &0.81\\
        64 &0.77 &0.82\\
        128 &0.75 &0.82\\
        256 &0.79 &0.80\\
    \end{tabular}
    \label{tab:width}
\end{table}
The gap is most prominent for 8 hidden features per layer, where DFA does not exceed the 10\% threshold of random guessing. Evidently, a certain degree of freedom (given by the number of weights) is required to align the weights and still have the flexibility to adjust them for learning.

\subsection{Impact of Circuit Imperfections on DFA-Based Training}
With the above findings in mind, the impacts of circuit parameters, namely ADC precision and subarray size, on network performance will be investigated. However, corresponding nonidealities cannot be integrated into the BP analysis. As BP involves complex data dependencies, analyzing circuit imperfections on the software side requires significant graphics memory. In contrast, DFA does not rely on maintaining its computational graph, and training analyses and assumptions can be tested. As both DFA and BP rely on the same hardware computations, general trends can be inferred for on-chip training. To cope with memory requirements, a three-layer network is preferred on the MNIST task for this analysis.

\begin{figure}[t]
    \centering
    \includegraphics[width=0.24\textwidth]{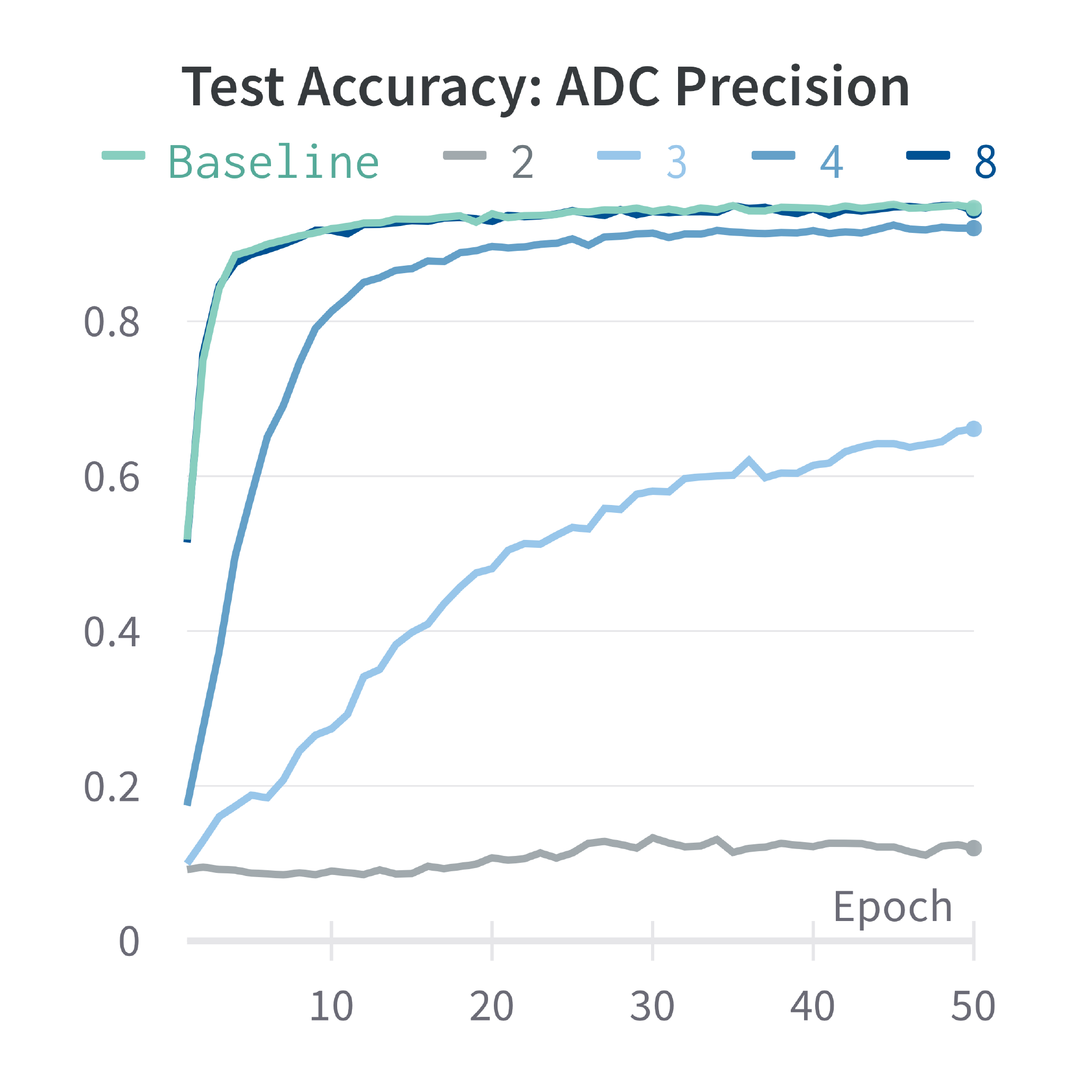}
    \includegraphics[width=0.24\textwidth]{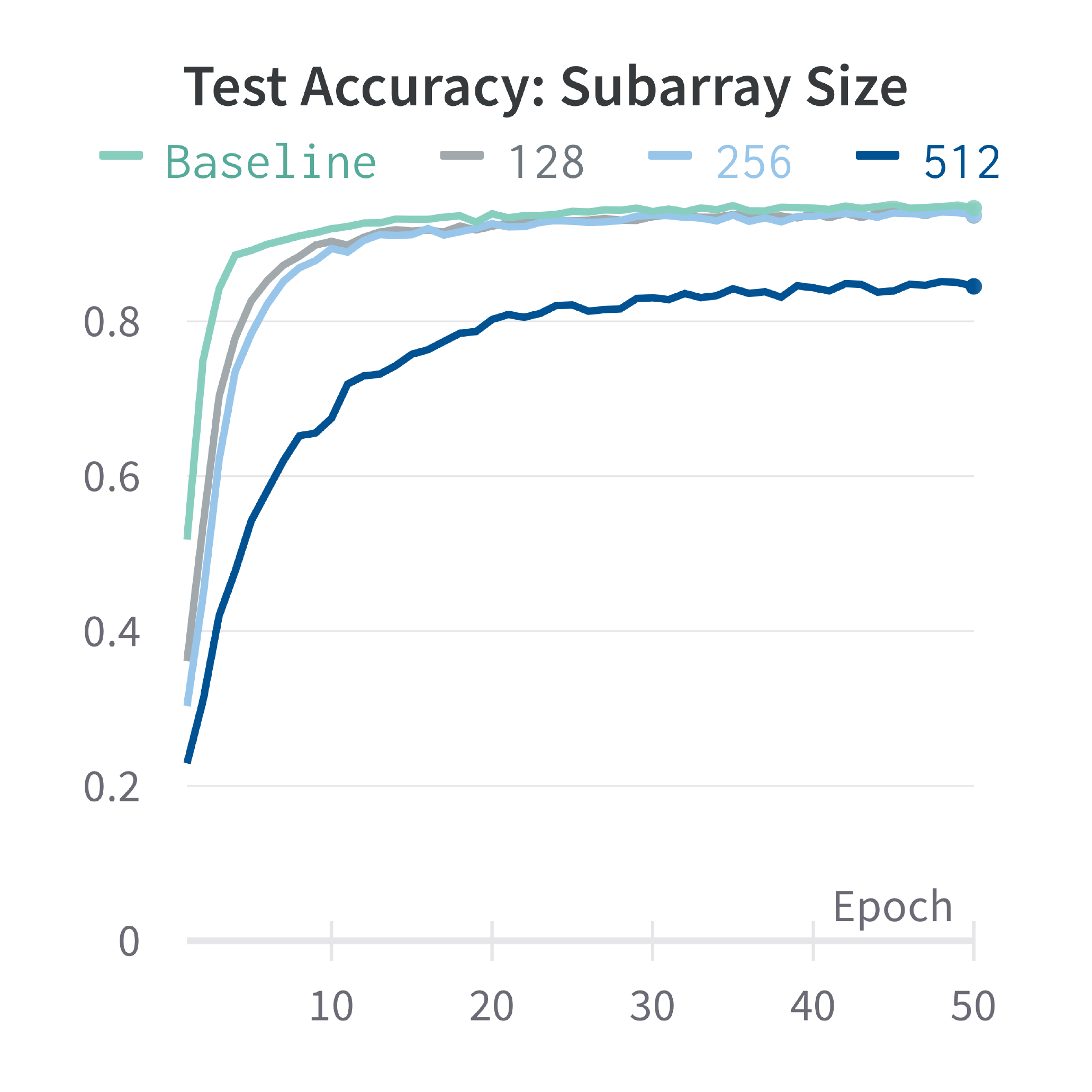}
    \caption{The effects of ADC precision (left) and subarray size (right) on DFA-based training. At least a 3-bit ADC is required for adequate accuracy results, while increasing subarray size increases errors due to IR drop.}
    \label{fig:circuitTrain}
\end{figure}
Fig.~\ref{fig:circuitTrain} displays how training stops for 1-bit ADC precision due to vanishing gradients and quantization errors, while the 2-bit precision does not achieve results above the 10\% threshold. In contrast, precision levels of 3 and 4 yield a significant improvement in accuracy. Still, a large number of epochs is required for convergence, indicating a slow learning process. At the same time, it should be considered that ADC precision has significant impact on area and energy consumption, constituting a major bottleneck in the CIM scheme. The tradeoff introduced by subarray size is linked to the overhead for peripheral circuits. Larger dimension subarrays invoke less overhead by processing more array data at each peripheral circuit. However, this also results in lower accuracy, as seen in Fig.~\ref{fig:circuitTrain}, due to the substantial IR drop associated with a larger subarray, generating noise in the partial sum accumulation.

\subsection{Noise Induced by Quantization \& Device Imperfections}
Unlike circuit nonidealities, device imperfections and predefined precision can be evaluated for both DFA and BP.
\subsubsection{Quantization Precision}
When considering quantization effects, the gradient vanishing problem becomes critical for BP, especially for deep networks. Sanfiz et al.~\cite{b8}, however, indicate that DFA is less sensitive to this because the errors are not propagated but passed directly to each layer. Still, it is assumed that with lower gradient precision, it becomes increasingly difficult to align the matrices $W$ and $B$. As a result, when sweeping weight and gradient precisions, learning for both algorithms stopped after 4 bits or less. Given a gradient precision of 5 bits, the error quantization leads to more pronounced differences. DFA learns at all swept precisions, although 1-bit began to improve only after the 50th epoch. For BP, learning is observed without delay.

\begin{figure}[b]
    \centering
    \includegraphics[width=0.45\textwidth]{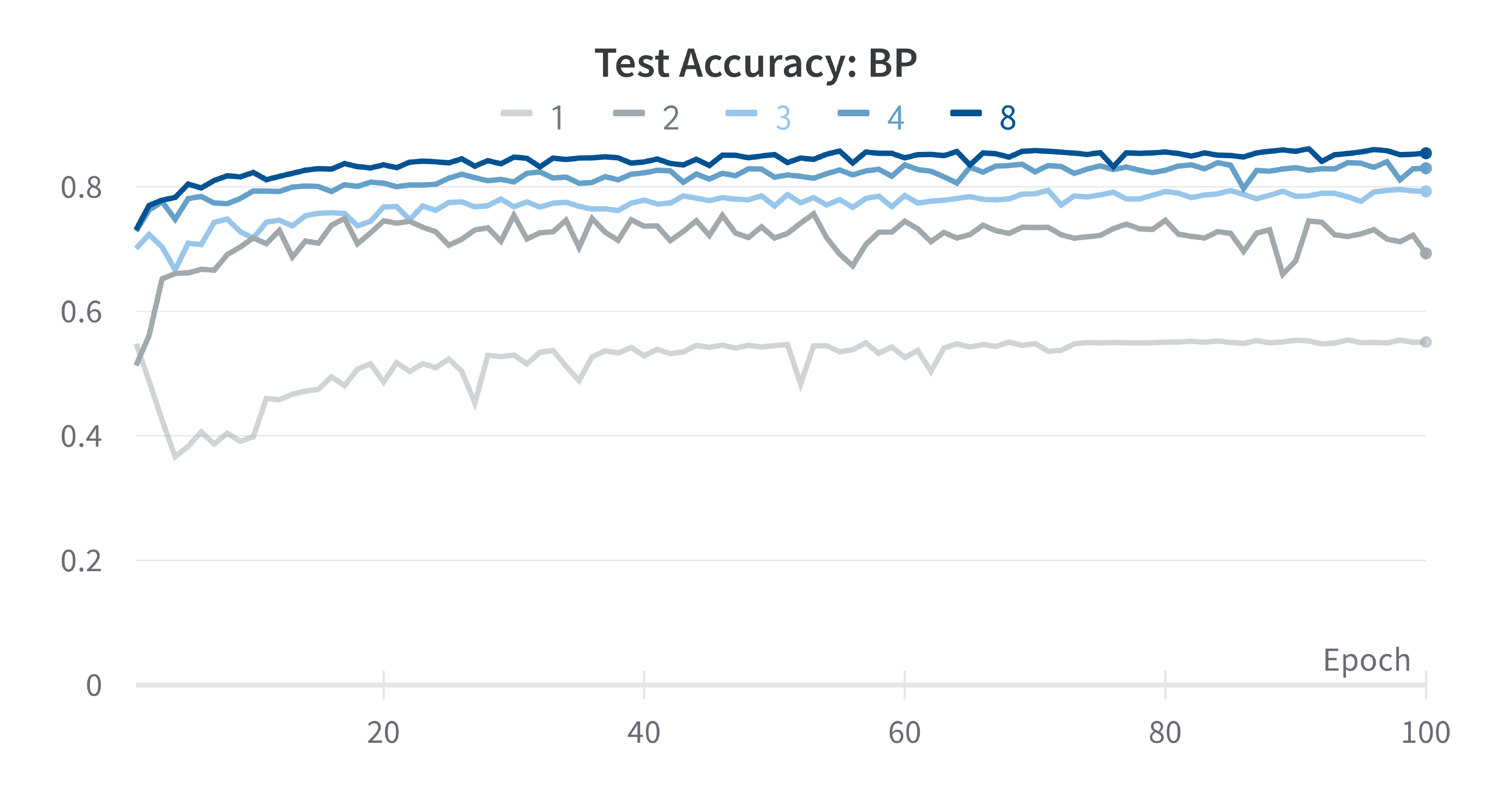}
    \caption{Effects of error precision on test accuracy for BP.}
    \label{fig:error}
\end{figure}
Focusing on the relative differences within a given learning rule in Fig.~\ref{fig:error}, however, visible performance steps are observed for BP for corresponding precision levels. In contrast, DFA yields almost the same accuracy for all precisions. One possible explanation is that DFA quantizes error only once before passing to each layer. For BP, on the other hand, the error is repeatedly quantized from layer to layer, resulting in a higher degree of accumulated error for initial layers at low precisions. Regarding hardware metrics, no substantial differences were observed for either learning rule in changing error precision due to the error's small dimensional size.

In addition, DFA is not observed having much robustness to general noise and quantization. In sweeping precision of the activation values, DFA only achieves competitive performance for at least 3 bits, whereas BP achieves high performance for all precisions tested. Since storing many activation values during training is essential, smaller precisions are preferred on the hardware side due to lower area, energy, and latency. Therefore, the quantization robustness of BP is a measurable advantage over DFA in this regard.

\subsubsection{Device Imperfections}
In addition to the precision of the models, hardware nonidealities can also lead to computational errors causing imprecise results. Experimentally demonstrated in this work, imperfections from device-to-device are absorbed during iterative training: No visible difference was observed across different levels of variability. However, error in programmed conductance states as a result of pulse-to-pulse (cycle-to-cycle) variation can have an impact of model performance. Sweeping typical ranges from 1\% to 5\% standard deviation over the mean of conductances, the learning rules yield fundamentally different results: BP converges to comparable accuracies over different degrees of variability, whereas DFA performs well only for 1\% deviation. More severe imperfections of 2\% lead to strongly fluctuating test accuracies with a drop in overall performance. 
As the cycle-to-cycle variation changes over time in contrast to the constant device-to-device variations, these errors are not absorbed during the iterative training of DFA. Furthermore, unlike BP, which is able to overcome this variation and converge, the cycle-to-cycle variation combined with the pressure for the forward weights to align with the random weights seems to prevent the algorithm from converging.

\subsection{Chip Area}
\begin{figure}[b]
    \centering
    \includegraphics[width=0.24\textwidth]{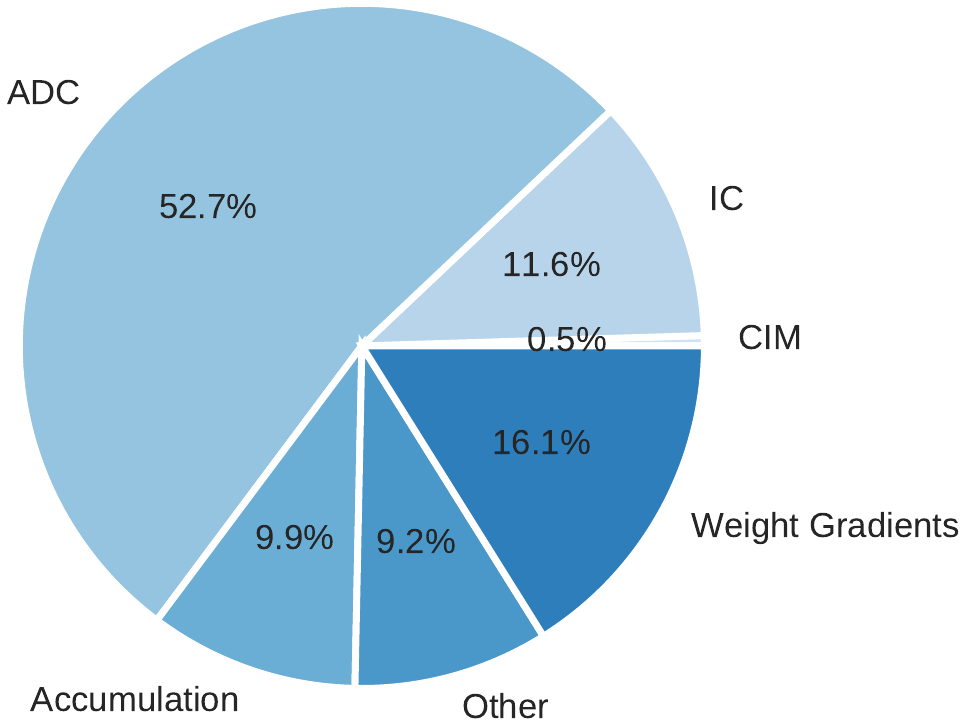}
    \includegraphics[width=0.21\textwidth]{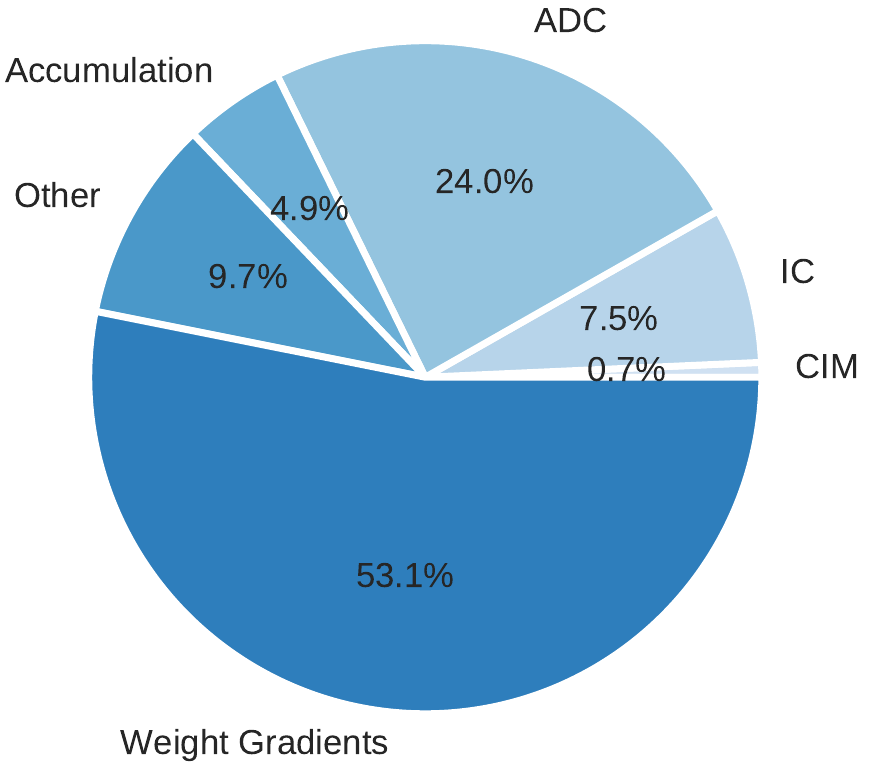}
    \caption{Breakdown of chip area for BP (left) and DFA (right). While the ADC circuitry consumes the most area for BP, the WGUs consume the most area for DFA.}
    \label{fig:areabreakdown}
\end{figure}

\begin{figure}
    \centering
    \includegraphics[width=0.49\textwidth]{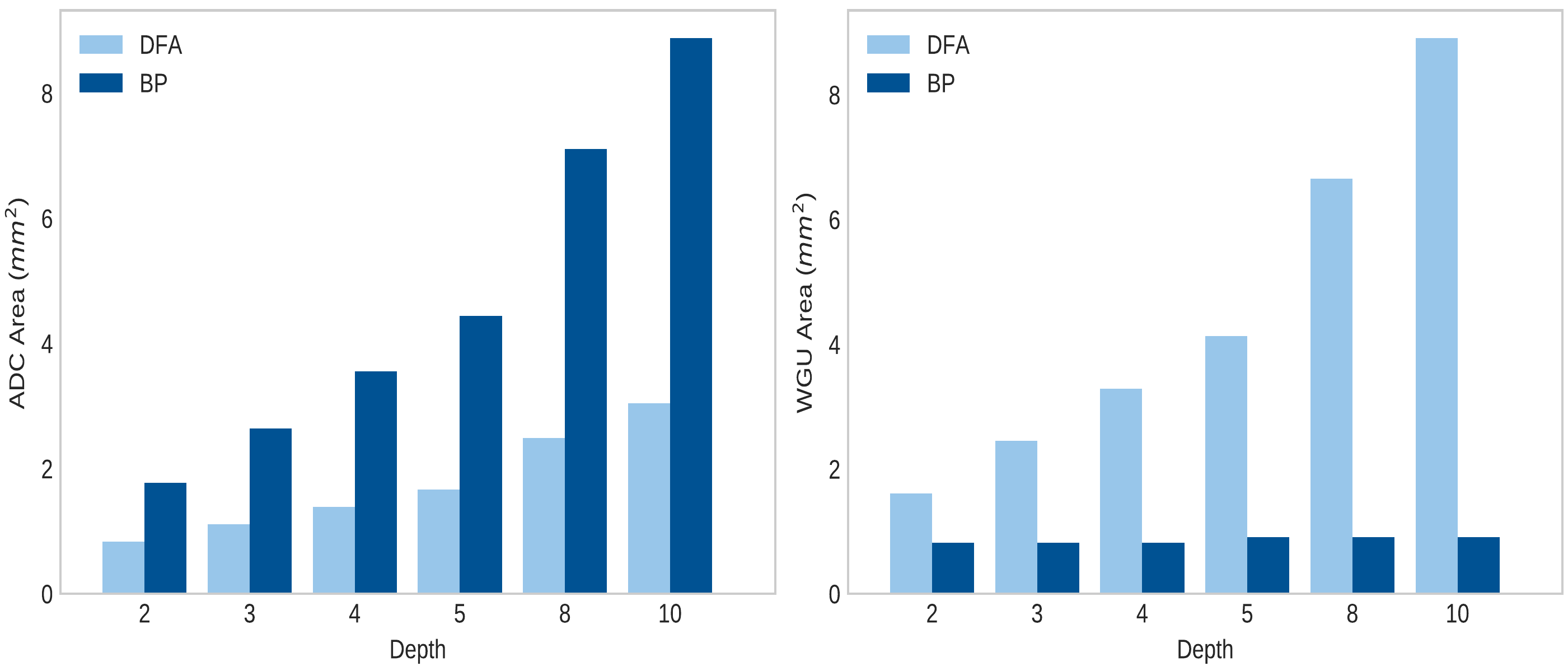}
    \caption{Hardware-specific area consumption over network depth, including ADC (left) and WGU (right) area for both BP and DFA.}
    \label{fig:growthrates}
\end{figure}

\begin{figure}[b]
    \centering
    \includegraphics[width=0.4\textwidth]{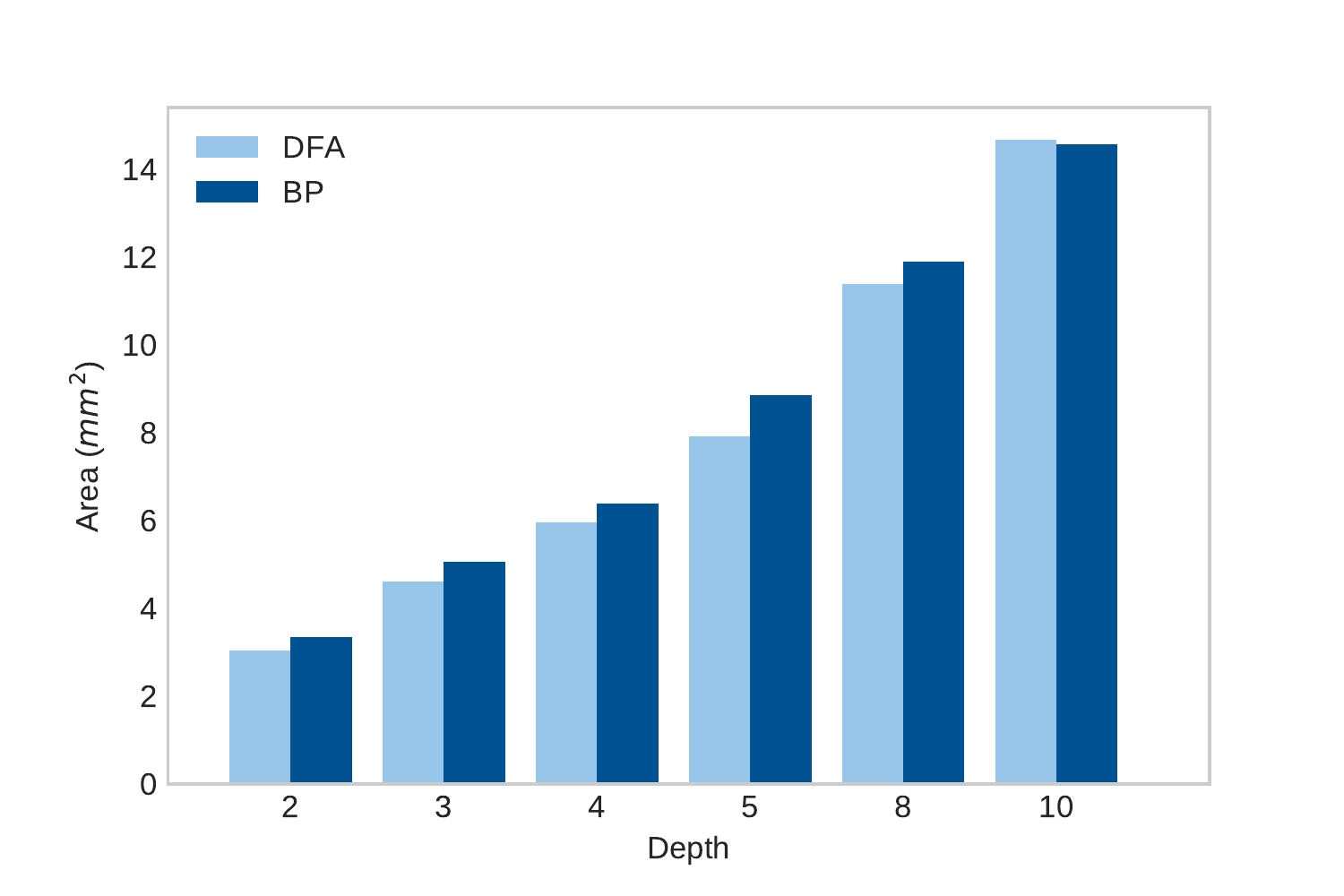}
    \caption{Total chip area for both BP and DFA over network depth.}
    \label{fig:area}
\end{figure}

As Fig.~\ref{fig:areabreakdown} shows, the additional weight matrix expressed by the CIM area consumption (representing the area in terms of memory cells) is marginal, consuming less than 1\% of total chip area.
Instead, peripheral circuits determine the majority of chip area, including ADCs, ICs (integrated circuits), accumulation logic, and weight gradient units (WGUs), with BP chip area mainly attributed to ADCs due to the need for transposable weights with twice the ADCs per CIM macro (left) and DFA chip area mainly attributed to WGUs due to parallel weight updates for each layer (right). Thus, the comparison between chip areas comes down to the driving forces behind the use of ADCs and WGUs, as well as how the number of both changes with deeper networks. Fig.~\ref{fig:growthrates} reveals the trends of both ADCs in BP and WGUs in DFA to be almost identical, resulting in similar trends for total chip area as summarized in Fig.~\ref{fig:area}.

\begin{figure}
    \centering
    \includegraphics[width=0.49\textwidth]{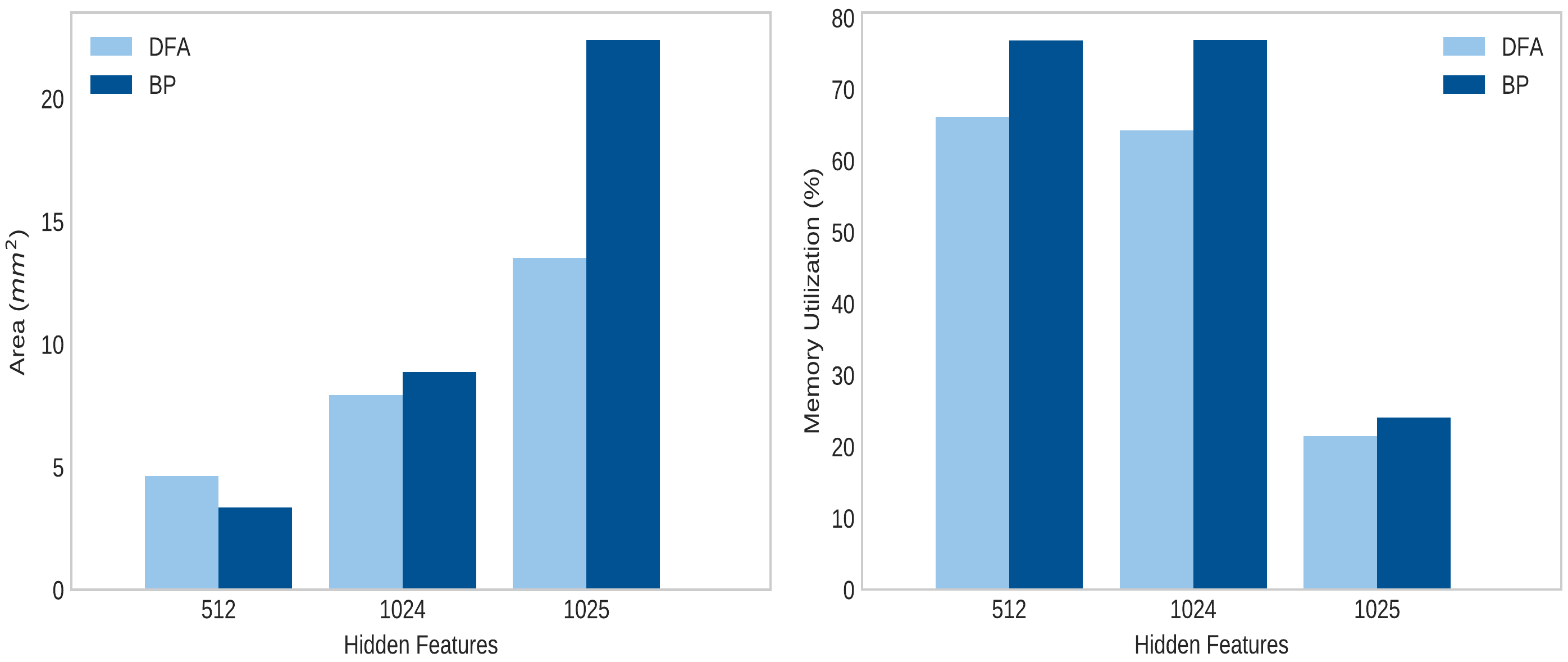}
    \caption{Impact of the number of hidden features on memory utilization (left) and chip area (right). A single tile's limit of 1024 memory cells means that 1025 hidden features requires more tiles, significantly increasing area and reducing memory utilization.}
    \label{fig:width}
\end{figure}

Despite the additional matrix and memory requirements, DFA consumes marginally less area than BP. As the number of layers increases, this saving becomes relatively smaller, resulting in DFA requiring slightly more area for 10 layers. 
Scaling with the network depth, WGUs implemented on SRAM cells become more costly in terms of area than the peripherals. Expanding the network in terms of matrix width, a more significant difference is observed for memory utilization: The discrete step of tiles becomes more evident when comparing the charts in Fig.~\ref{fig:width}, where for a fixed network size of 5 layers the number of hidden features is altered. At 1025 hidden features, the tile limit of 1024 memory cells is exceeded. Therefore, a 1x1 tile structure has to be extended to a 2x2 structure to account for all weights. This results in low utilization and a significant area overhead for peripherals.

\subsection{Energy Consumption}
Previous sections discussed a reduction in the number of peripherals and FLOPs required to compute the gradients in DFA. Breaking down energy consumption into different hardware categories, however, shows a 95\% share is allocated to accessing off-chip buffers. Activations, errors, and gradients have to be loaded to and from the off-chip memory, making buffering the bottleneck of the training process. Hence, off-chip accesses become the driving force of energy consumption, underlining the importance of the CIM paradigm. Even though FLOPs are saved, the resulting matrices are of the same size, and so are the intermediate values. Thus, no differences in buffer accesses are observed for DFA and BP, resulting in virtually the same energy consumption.

\subsection{Training Latency}
\begin{figure}[t]
    \centering
    \includegraphics[width=0.4\textwidth]{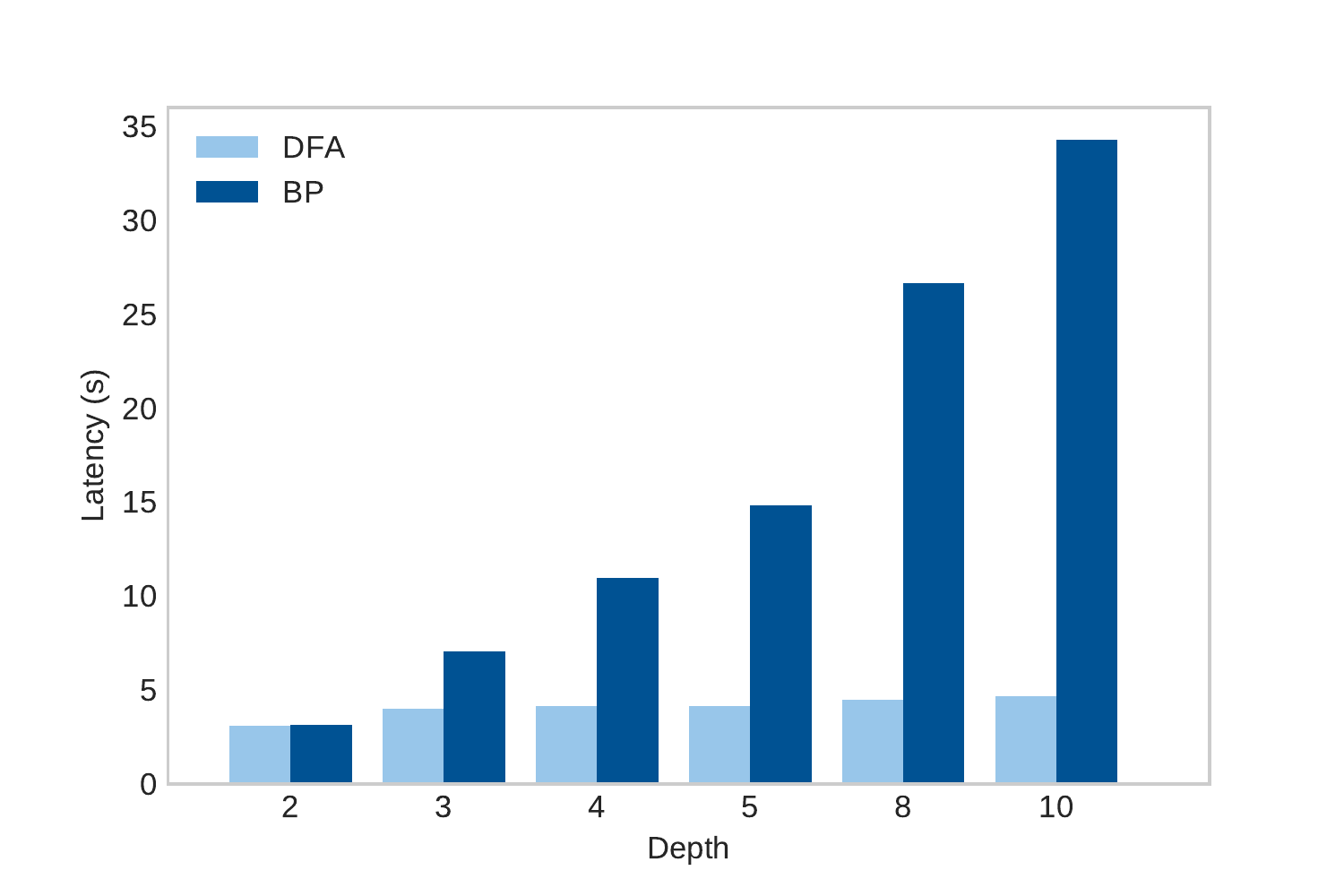}
    \caption{Training latency for BP and DFA over network depth. The ability of DFA to parallelize weight updates leads to considerable reduction in latency.}
    \label{fig:latency}
\end{figure}

A similar pattern is seen for training latency, where 91\% thereof is not caused by the actual computations but by buffering intermediate results. Given the assumption that different layers access different DRAM blocks in parallel, a considerable speedup can be achieved by parallelizing the weight gradient calculation and writing, which represent 96\% of the overall latency. Thus, fundamental differences can be observed when compared over network depth, as shown in Fig.~\ref{fig:latency}.

In contrast to BP with strictly sequential computational flow, DFA is executed in a heavily parallelized scheme. As network depth increases, the latency for BP grows linearly, while DFA takes almost the same training time for an increased number of layers. As more layers are added, the ratio of latencies of BP to DFA approaches a scaling factor of $N$ for $N$ parallelized layers. Considering the central finding of subsection A that DFA requires more epochs to converge, the speedup with respect to performance is reduced, however. Still, parallelization enables substantial acceleration and thus higher throughput.

\section{Conclusion}
While Backpropagation yields better model performance in most cases, especially in noisy environments, Direct Feedback Alignment resolves the bottleneck of sequential dependencies, thus allowing for parallelization. In resource-constrained environments, such as edge devices, DFA provides a valuable alternative: Tradeoffs between area and latency due to incorporating additional hardware can be adjusted for each use case. When a smaller chip design is preferred, the degree of parallelization and the number of area-costly WGUs can be decreased. This potentially saved area could be utilized to increase the precision of quantized values, e.g., activation values. Hence, DFA could allow for higher precision values at the same chip area as BP. Accuracy improvements would be achieved, allowing DFA to overcome its imprecise nature.

From a biological perspective, DFA on neuromorphic hardware enables local computations and therefore solves fundamental implausibilities of learning in artificial neural networks. Spiking neural networks could build on this and further enhance the idea of bio-plausibility. This would allow for even higher energy efficiency and noise tolerance, which have been discussed for conventional neural networks in detail in this paper.

\bibliographystyle{plain}
\bibliography{references}

\end{document}